\documentclass[twocolumn]{article} 
\usepackage{graphicx} 
\usepackage{authblk} 
\usepackage{natbib} 
\usepackage{hyperref}
\usepackage{multirow}
\usepackage{amsmath}

\title{Adaptable Embeddings Network (AEN)}

\author[1]{Stan Loosmore\thanks{loosmore@usc.edu}}
\author[1,2]{Alexander J. Titus\thanks{publications@theinvivogroup.com}}
\affil[1]{University of Southern California}
\affil[2]{In Vivo Group}
\date{Nov 2024}

\begin{document}

\maketitle

\begin{abstract}

Modern day Language Models see extensive use in text classification, yet this comes at significant computational cost. Compute-effective classification models are needed for low-resource environments, most notably on edge devices. We introduce Adaptable Embeddings Networks (AEN), a novel dual-encoder architecture using Kernel Density Estimation (KDE). This architecture allows for runtime adaptation of classification criteria without retraining and is non-autoregressive. Through thorough synthetic data experimentation, we demonstrate our model outputs comparable and in certain cases superior results to that of autoregressive models an order of magnitude larger than AEN's size. The architecture's ability to preprocess and cache condition embeddings makes it ideal for edge computing applications and real-time monitoring systems. \end{abstract}

\section{Introduction}
Many modern-day AI tasks utilize auto regressive language models (LMs), most notably, Generative Pretrained Transformers (GPT) have taken the world by storm(\cite{naveed2024comprehensiveoverviewlargelanguage}). LMs are notable for their flexibility, often requiring minimal fine-tuning for task adaptation. Additionally, their ease of access, facilitated by open-source availability and through API services, makes them practical for widespread use (\cite{brown2020languagemodelsfewshotlearners}). 

LMs demonstrate strength in text classification tasks, determining if input text corresponds to specific labels (\cite{lepagnol2024smalllanguagemodelsgood}). For GPT models, this typically involves evaluating the text through a single prompt combing both classification requirements and said text within a single prompt, as seen in applications like sentiment analysis \cite{kheiri2023sentimentgptexploitinggptadvanced}. A notable subset of these classification tasks focus on prompt engineering detecting specific entities or topics within text also known as Name Entity Recognition or NER \cite{wang2023gptnernamedentityrecognition}. Consider, for instance, analyzing a corpus to determine whether it contains references to "Thermodynamics." This application requires two primary components: the target corpus for analysis and the evaluation criterion, which takes the form of a binary (yes/no) assessment regarding the presence of a particular topic or concept. 

 The evaluation criteria or 'natural language if statement' becomes particularly useful when examining constantly updating data, such as a real-time transcript. In such scenarios, it's valuable to continuously evaluate the incoming text against specific criteria. \newline For example, in a live speech transcription:

We might want to know if the speaker has mentioned a key topic.
We could check if certain commitments or promises have been made.
We may need to flag when sensitive information is discussed.

At each update of the transcript, an ideal model generates boolean values indicating satisfaction of specific conditions. An efficient classifier enables real-time monitoring beyond speech analysis. Modern LMs used for this task often incur high computational costs. This makes it impractical for resource-constrained, quick-paced environments like edge computing, mobile devices, or environments requiring many criteria to be evaluated simultaneously. 

Embedding models, especially transformer-based ones, have demonstrated superior performance in classification tasks, offering both high accuracy and significantly greater efficiency compared to decoder-based transformers at lower parameter counts (\cite{tunstall2022efficientfewshotlearningprompts}). Decoder-based transformers are not the only models capable of capturing semantic relationships dynamically. Embedding architectures produce vector representations of complex, non-numeric data such as text or images, capturing the semantic relationships within the data. Sentence Transformers, a type of attention based embedding model, typically achieve this by mean pooling the representations of attended tokens from a fine-tuned Bidirectional Encoder Representations from Transformers (BERT) model (\cite{devlin2019bertpretrainingdeepbidirectional, reimers2019sentencebertsentenceembeddingsusing}). Recent methods, such as transforming pre-trained auto-regressive LMs to embedding models, have also shown promise (\cite{bauer2024comprehensiveexplorationsyntheticdata}).

Embedding proves usefulness through relativity. The similarity between embeddings can be mathematically quantified using techniques such as Cosine Similarity or Euclidean distance, producing a continuous value that represents their degree of likeness \cite{levy2024guidesimilaritymeasures}. Embedding models have seen less use in classification as training involves producing continuous scores rather than discreet one. Training most typically uses contrastive loss , clustering similar labeled embeddings close to each other, or regressive loss aimed at approximating a similarity score (\cite{reimers2019sentencebertsentenceembeddingsusing}, \cite{bauer2024comprehensiveexplorationsyntheticdata}). Additionally, traditional feed-forward neural networks, which typically rely on fixed input-output mappings, lack capacity to use a "prompt" to classify inputs. They do not inherently possess the mechanisms to adapt \textit{what} they are classifying or account for nuanced differences in input text post training, making them less suitable for tasks that require a flexible, context-aware approach (\cite{Goodfellow-et-al-2016})

To address these challenges, we introduce Adaptable Embeddings Networks (AEN), a novel and computationally efficient method for classifying text based on natural language criteria specified at runtime. AEN leverages a dual-encoder architecture: A query encoder for processing the input text, and a criterion encoder for interpreting the classification rules.
These encoders feed into a hierarchical networks combining both inputs to create a highly adaptable classifier. Our approach significantly outperforms leading Small Language Models (SLMs) in efficiency, often by orders of magnitude, while maintaining comparable accuracy in binary text classification tasks.

This paper presents a novel approach to network architecture and efficient prompting for resource-constrained environments.

\section{Related Work} 
In this section we give a brief overview of embedding techniques, most notably Sentence Transformers. Then we discuss synthetic and augmented data as a means of generating additional data when little to none exists.    
\subsection{Embedding Techniques}

\subsubsection{Training Methods}
After obtaining a strong pre-trained LM, comes the process of fine-tuning to produce meaningful embeddings. Prominent fine-tuning approaches/network architectures include Siamese, Triplet, Cross Encoders \cite{reimers2019sentencebertsentenceembeddingsusing}.

\textbf{Siamese Networks (Bi-encoders)}: These networks, also referred to as bi-encoders, use two mirrored encoders joined by one or more linear layers to predict the similarity between inputs, generally producing a score between zero and one. The encoder is fine-tuned based on this similarity measure (\cite{conneau2018supervisedlearninguniversalsentence}).

\vspace{0.3cm}
\textbf{Triplet Networks}: This architecture employs three identical encoders processing three distinct text inputs: two semantically similar texts and one dissimilar text. The network learns by minimizing the distance between similar text embeddings while maximizing the distance between dissimilar pairs, enabling more nuanced differentiation in the latent space (\cite{ein-dor-etal-2018-learning, Schroff_2015}).

\vspace{0.3cm}
\textbf{Cross Encoders}: This network utilizes a single encoder to process two concatenated sentences simultaneously (\cite{devlin2019bertpretrainingdeepbidirectional}). While Cross Encoders capture the relationships between the two inputs effectively, they are unable to generate individual embeddings (\cite{reimers2019sentencebertsentenceembeddingsusing}). If used with attention based these models incur additional computational cost due to its quadratic complexity.

\subsubsection{Embedding with Transformers}
The Universal Sentence Encoder was first embedding model to utilize the transformer architecture (\cite{cer2018universalsentenceencoder}). More recently, \cite{reimers2019sentencebertsentenceembeddingsusing} proposed
Sentence Transformers, an extension of the Bidirectional Encoder Representations from Transformers, (BERT) architecture. Base BERT trains transformers on predicting hidden or masked words, and next sentence prediction; one sentence logically following another \cite{devlin2019bertpretrainingdeepbidirectional}. BERT performs well on these task. However BERT, lacks the ability to encode meaningful fixed length representations from a variable input length, motivating (\cite{reimers2019sentencebertsentenceembeddingsusing}) to create Sentence BERT (SBERT). Created through tuning BERT using triplet and siamese architectures, SBERT produced meaningful embeddings. 

\subsection{Low Data Environments}
To overcome the challenges presented by environments with low data two main techniques: augmentation and synthetic data creation. 

\subsubsection{Augmented Data}
Augmenting data is the practice of manipulating existing data creating additional high quality data (\cite{wang2024comprehensivesurveydataaugmentation}). Using augmentation requires a existing unlabeled data.
When a small amount of labeled data exists, it is generally used to train a labeling model for the rest of the data. In the case of SBERT researches developed Augmented SBERT (AugSBERT). If labeled "gold" data is available a cross-encoder model is fine-tuned on this high-quality labeled data. The fine-tuned cross-encoder is then used to label additional unlabeled data.
If no labeled data is available one would use a pre-trained model to label the data. For AugSBERT they present a pre-trained cross-encoder model used directly to label unlabeled data. This labeled data is then used to train downstream \cite{thakur-2020-AugSBERT}).

\subsubsection{Synthetic Data}
Synthetic Data, commonly referring to data produced by a model or algorithm rather then from observation or humans (\cite{bauer2024comprehensiveexplorationsyntheticdata},) is employed when little to no data exsist unlabeled or otherwise. In particular synthetic data has shown promise in training language models (\cite{li2023textbooksneediiphi15}). Companies have also used synthetic data to adjust embeddings to reflect the world more accurately (\cite{jina2024synthetic}).

\subsection{Kernel Density Estimation (KDE)}
\begin{figure}[t]
\centering
\includegraphics[width=.5\textwidth]{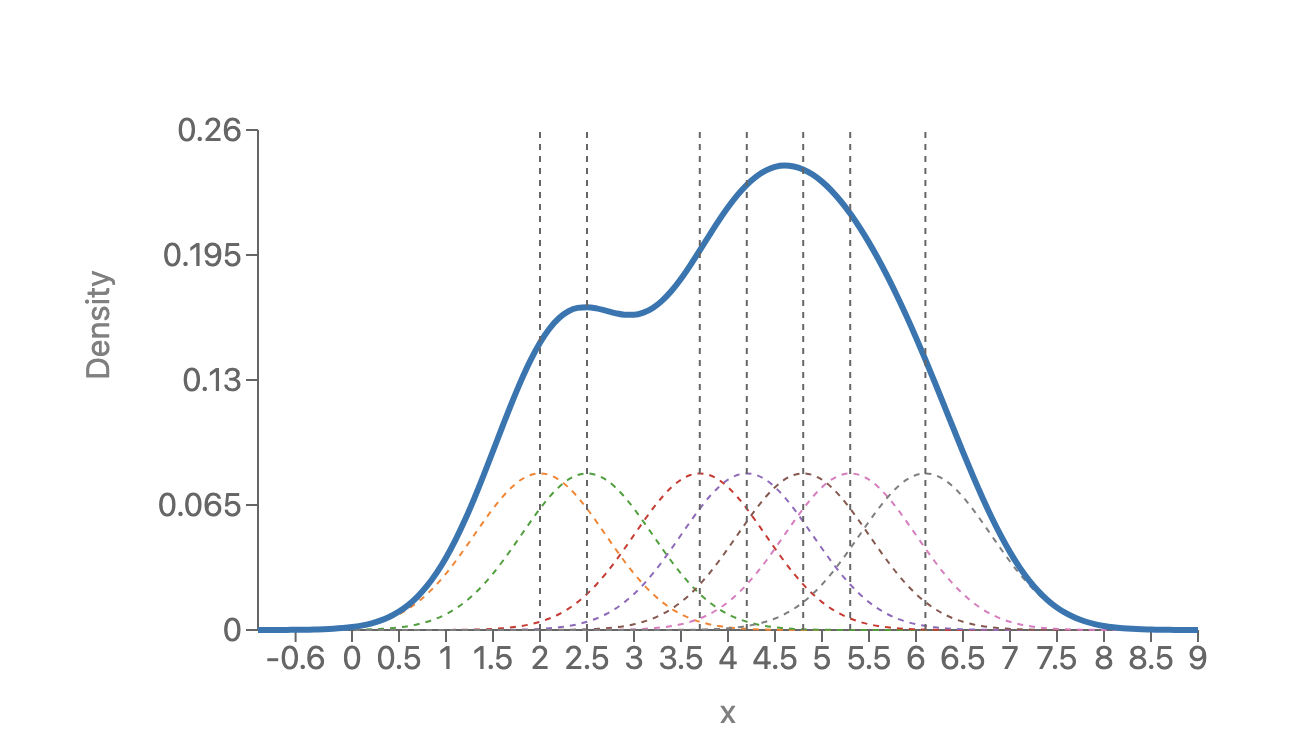}
\caption{1D Kernel Density Estimation showing individual kernels (dashed lines) and their sum (solid line)}
\label{fig:KDE example}
\end{figure}

Kernel Density Estimation (KDE) is one of the most widely regarded methods for estimating probability distributions from datasets. Its popularity arises from its ability to create smooth, continuous probability density estimates without assuming a specific parametric form (like normal or exponential) \cite{chen2017tutorialkerneldensityestimation}. KDE leverages existing data points by placing a kernel function at each observation and summing these kernels to construct a probability density function, as illustrated in Figure~\ref{fig:KDE example}.
The general form of a kernel density estimator is given by Equation~\ref{eq:kde_general}:
\begin{equation}
\label{eq:kde_general}
\hat{f}(x) = \frac{1}{nh}\sum_{i=1}^{n}K\left(\frac{x-x_i}{h}\right)
\end{equation}
where $\hat{f}(x)$ is the estimated density at point $x$,
$n$ is the number of observations in the dataset,
$h$ is the bandwidth (smoothing parameter),
$K(\cdot)$ is the kernel function, and
$x_i$ are the individual data points.
The kernel function $K(\cdot)$ is typically a symmetric probability density function, with the Gaussian kernel being a common choice, as shown in Figure~\ref{fig:KDE example}.
The Gaussian kernel is defined in Equation~\ref{eq:gaussian_kernel}:
\begin{equation}
\label{eq:gaussian_kernel}
K(u) = \frac{1}{\sqrt{2\pi}}e^{-\frac{u^2}{2}}
\end{equation}
Several methods exist for bandwidth estimation, with Scott and Silverman's notable approaches calculating bandwidth as a function of both the standard deviation and the number of datapoints used to create the KDE (\cite{wells2017simpleefficientdensityestimator}). The choice of bandwidth, significantly affects the smoothness of the resulting density estimate.

\section{Methods}

In this section we first discuss our data gathering and generation methods, then our training process and Model Architectures. 

\subsection{Data}

Our research requires a specific type of format: 
a statement, and condition, with a binary label. We index specifically for a transcript like data where if one had access to the transcript they could use it to perform actions. This section details how and why we gather data of this format. 

\subsubsection{Motivation for Synthetic Data}
To our knowledge, no existing datasets combine natural language statements and classification criteria in the specific format required for this task. While existing conversational datasets could theoretically be adapted through human annotation, the scale required for effective model training would make this approach prohibitively resource-intensive. Furthermore, using large language models to transform existing datasets would likely introduce similar biases and artifacts as our synthetic generation approach, while adding unnecessary complexity to the pipeline. Therefore, we opted for direct synthetic data generation, allowing us to precisely control the data distribution and maintain consistent evaluation criteria throughout our experiments.

We generate statements, conditions, and labels synthetically through a multi-step pipeline utilizing OpenAI's batch API using GPT 4o-Mini unless otherwise specified. This approach offers several advantages:

\begin{enumerate}
    \item Control over syntax and format: Synthetic generation allows us to shape the linguistic structure and format of the data precisely to our needs, ensuring consistency across data points for ideal manipulation.
    \item Scalability: We can generate large volumes of data quickly and in a cost effective manner, essential for robust pre-processing.
    \item Customization: We can tune the generation pipeline to cover a varying scenarios and edge cases potentially underrepresented in real-world datasets.
    \item Ethical considerations: Synthetic data mitigates privacy concerns that might arise from using real-world conversational data.
\end{enumerate}

Synthetic data also comes with drawbacks. Studies show LM inaccurately represent real world data through presenting unrealistic scenarios and additional bias incurrent in training \cite{veselovsky2023generatingfaithfulsyntheticdata}. We tackle this issue through employing robust prompting and format cleaning methods.

\subsubsection{Data Generation Process}
\begin{figure}[h!]
    \centering
    \includegraphics[width=0.3\textwidth]{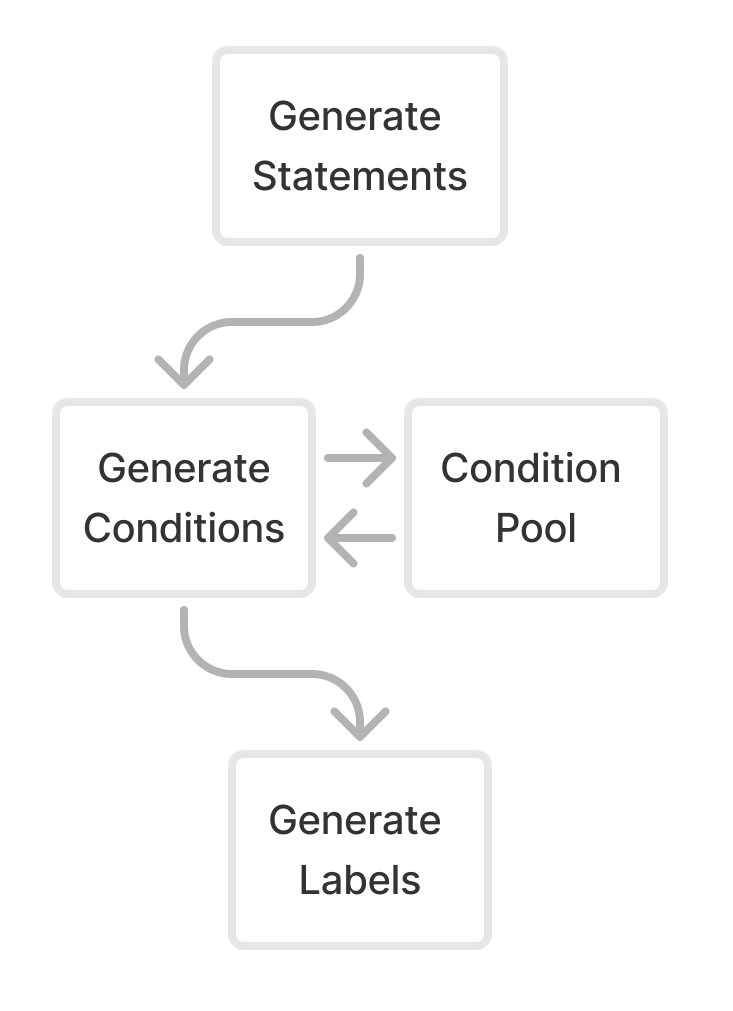}
    \caption{Data Generation Process}
    \label{fig:data_line}
\end{figure}

Generating synthetic data through LMs presents a significant challenge: ensuring both data diversity and maintaining coherence (\cite{veselovsky2023generatingfaithfulsyntheticdata}). In our approach, we prioritized diversity of potential transcripts while implementing controlled patterns that allow us to prove convergence on data meeting our defined requirements. These patterns are further detailed below. We leave the expansion to broader domains of data for future work.

We aimed to generate diverse data while focusing on a defined information subset exhibiting repeatable patterns: contextual and linguistic. This strategy allowed us to focus converging on a manageable set of data that meets our defined requirements, rather than attempting to cover the entire spectrum of possible outputs. 

We implemented a multi-shot approach with carefully engineered prompts to optimize data generation (examples can be found in Appendix~\ref{app:prompts}). By investing significant effort in prompt design and providing plausible examples to the model, we guided outputs toward our desired format and content. This methodology maximized the efficiency of API calls by enabling us to utilize all parsable outputs from each batch, making the most of our computational resources.

\paragraph{Statement Generation Approach}

The final data must contain labeled statement and condition pairs. The challenge lies in ensuring a balanced distribution of labels. 

We considered two potential approaches:

\begin{enumerate}
    \item Generate statements and conditions separately, then attempt to match them.
    \item Generate statements first, then use them to inform the generation of appropriate conditions.
\end{enumerate}

We found that generating them sequentially yielded a class imbalance of 1 to 6 compared a significantly higher proportion then in parallel. Therefore we choose generating the data sequentially. This approach allows us to use the content and context of each statement informing the generation of the condition. By doing so, we not only increase the likelihood of creating logically consistent and meaningful statement-condition pairs but also maintain a more balanced dataset suitable for effective model training.

\paragraph{Statement Generation Implementation}

We encouraged data variety through creating a prompt template taking in two inputs:  

\begin{quotation}
``Generate a conversation 25 statements long set in the context of \texttt{\{context\}}. During the conversation someone should \texttt{\{action\_data\}} and perform any other relevant software actions.''
\end{quotation}

\textbf{Context Generation:} The \{context\} input comes from a list of thousands of different settings in which the conversation may occur. We create this list through transforming a subset of Wikipedia Articles into potential settings/enviroments through OpenAI GPT calls. 
Here we take inspiration from (\cite{bauer2024comprehensiveexplorationsyntheticdata}) prompting:
\begin{quotation}
``What group activity \texttt{\{random.choice(verb\_list)\}} the noun? Respond with one in 5 or less tokens. Respond explicitly (don't use the word it).''
\end{quotation}

\textbf{Action Data Generation:} The \{action\_data\}, a list of potential actions, is derived from a similar process. Instead of Wikipedia Articles we use the top 1000 websites by traffic, picking actions that can be performed on each of these websites. We include actions in prompt because the primary use case through conversation is to monitor if someone expresses interest in an action. However, we create non-action oriented conditions for more generalizability; more about this in the next section.

This approach leads to tens of millions of possible prompts for statement generation. Additionally, we maintain a set of slightly differently worded instruction prompts and a set of curated examples to ground the model for more real-world results. We decide to generate whole conversations (25+ statements per call) so we can reap the benefits of a developing story rather than individual statements from a speaker. We found that generating one off statements/ paragraphs from GPT calls limited the diversity of responses. The condensed form provides additional benefits when creating statements.

For the Language Model settings at the statement generation stage, we use a temperature of 1.6 and a Top P of 0.85.
These settings encourage creativity and diversity in the generated statements while maintaining a reasonable level of coherence.

\paragraph{Condition Generation}

Following the generation of statements, we proceed to create conditions that these statements might satisfy.

Our condition generation process follows these steps:

\begin{enumerate}
    \item \textbf{Transcript-Specific Condition Generation:}
    We use each transcript (a set of 25+ statements) to inform the generation of a set of conditions (3+). By making a separate condition API call for each transcript, we virtually guarantee  every transcript will contain statements that satisfy the generated conditions. 

    \item \textbf{Database Storage:}
    We save every generated condition to a database. This allows us to maintain a diverse pool of conditions for continued use and analysis.

    \item \textbf{Cross-Pollination:}
    To increase data diversity, we cross-pollinate transcript-specific conditions with other transcript data groups. This means that conditions generated for one transcript are also paired with statements from other transcripts. This process helps to:
    \begin{itemize}
        \item Increase the variety of statement-condition pairs
        \item Reduce potential biases arising from always pairing conditions with their original transcripts
        \item Introduce comparisons between topics other wise unrelated. We allow sparse parts of the vector space to interact with each other increasing model robustness.
    \end{itemize}

\end{enumerate}

For the condition generation stage, we set the Language Model parameters as follows: a temperature of 1 and a Top P of 1. These settings allow for creativity in condition generation while maintaining coherence. The higher temperature compared to statement generation (1 vs 1.6) ensures more focused conditions, crucial for the desired evaluation specifications.

Our approach to condition generation complements our statement generation process, resulting in a dataset of statement-condition pairs suitable for training and evaluation.

\paragraph{Labeling Generation}

The final step in our data generation pipeline involves labeling the statement-condition pairs. A transcript comes with at least 75 (25*3) distinct statement condition pairs . This process presented unique challenges:

\begin{itemize}
    \item Large language models GPT-4o and its mini version struggled with labeling entire transcript datasets efficiently.
    \item Labeling each statement condition pair proved cost-inefficient.
\end{itemize}

To address these challenges, we developed the following approach:

\begin{enumerate}
    \item We pair each statement with the full set of generated conditions.
    \item For each statement, we make a single API call to label it against all conditions.
    \item This method balances efficiency and cost-effectiveness, allowing us to generate a comprehensive set of labels without excessive API usage.
\end{enumerate}

For the labeling process, we set the Language Model parameters as follows: a Temperature of 0
and a Top P of 1.
These settings prioritize deterministic outputs, ensuring consistency in the labeling process. The low temperature (0) minimizes randomness, needed for reliable binary classification.

This labeling strategy completes our data generation pipeline striking the balance between diversity and determinism.

\subsection{Model and Training}
\begin{figure}[t]
  \centering
  \includegraphics[width=.4\textwidth]{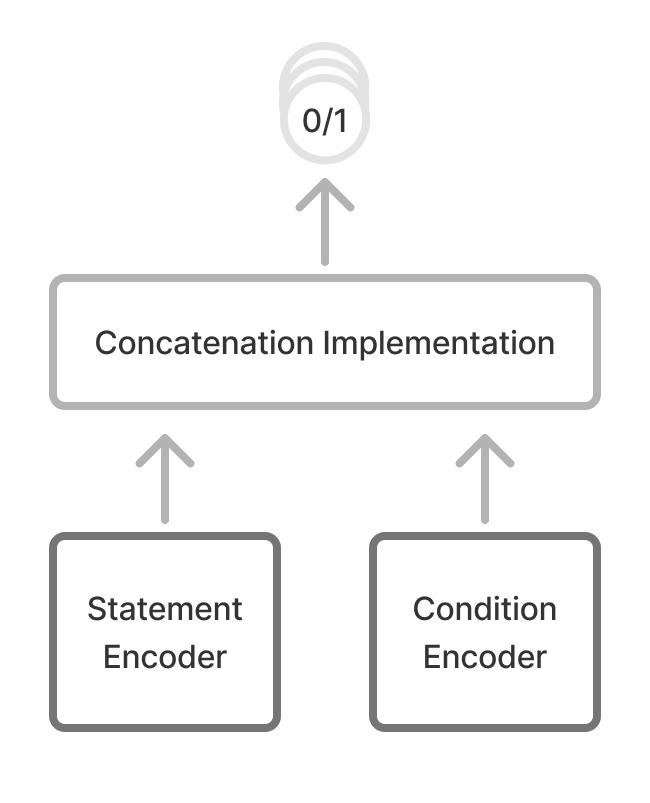}
  \caption{AEN General training architecture}
  \label{fig:train}
\end{figure}

\begin{figure}[t]
  \centering
  \includegraphics[width=.4\textwidth]{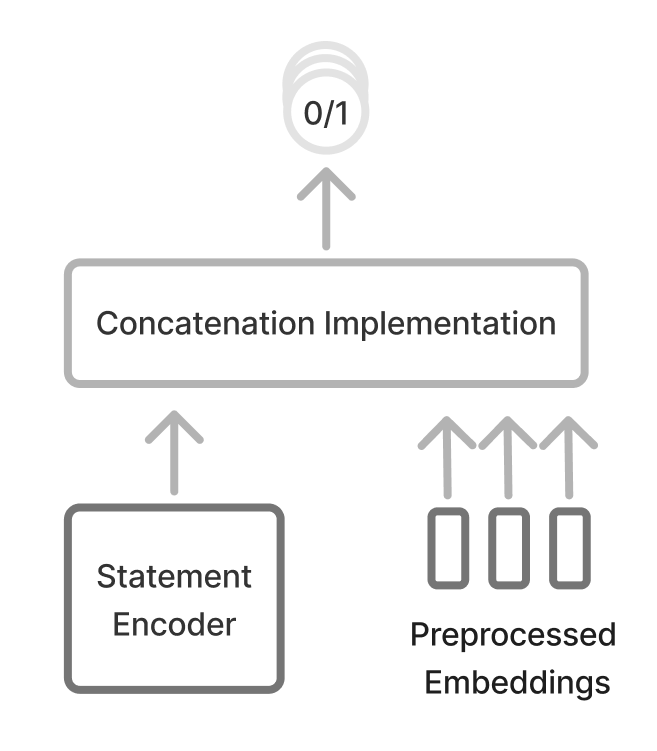}
  \caption{AEN runtime architecture for constant data evaluation}
  \label{fig:run}
\end{figure}

In this section, we outline our training and modeling methods culminating in the Adaptable Embeddings Networks (AEN), a compute-efficient zero-shot classifier designed to determine if a given text meets specified semantic criteria. AEN utilizes a dual-encoder architecture: one for processing input text and another for interpreting classification rules. We opted for a non-mirrored Siamese architecture with a binary classification head for training (Figure \ref{fig:train}), as it better suits our binary classification task compared to alternatives like cross-encoders or triplet networks \cite{reimers2019sentencebertsentenceembeddingsusing, Schroff_2015}. This approach allows for preprocessing of conditions and single-pass processing of new statements, optimizing computational efficiency for real-time applications (Figure \ref{fig:run}). The following section details our methodology for processing embedding vectors to produce classifications.

\subsubsection{Data Preparation}
Our data preparation process involves two main steps:
\paragraph{Batching Strategy}
To optimize processing efficiency, we batch each statement by token length. This approach minimizes the total padding required, ensuring that statements with similar padding are grouped within the same batch.
\paragraph{Condition Preprocessing}
Every condition in our dataset initially starts with the phrase "When someone". This standardization was helpful for GPT when it was labeling the statement-condition pairs. However, for our model input, we remove these two words from every condition.
This streamlined preparation process ensures our data is consistently formatted and optimized for our model architecture.

\subsubsection{Training Methodology}
We train on a NVIDIA RTX 4090. Data points, Learning Rate and Batch size, all varied from 50k to 5M points, 1e-5 to 1e-6 and 64 to 256 respectively. 
We employ the all-MiniLM-L6-v2 and all-mpnet-base-v2 models from the SBERT library on Hugging Face, based on the work by \cite{reimers2019sentencebertsentenceembeddingsusing}. These models are trained use mean pooling so we primarily adopt the same operation, however make some changes depending on the concatenation method.

\subsubsection{Exploration of Model Architectures}

We explored three main model routing architectures. This section provides a qualitative discussion of each approach, culminating in our final AEN model.

\paragraph{Feed Forward Networks}
Our initial approach involved generating separate embeddings from the encoders and using these as inputs to a Neural Network (NN). The input vector shape consisted of some factor of the output embedding vector length. For outputs $U$ and $V$ from each encoder, we tested various input combinations: $(U, V)$, $(U, V, |U-V|)$, $(U, V, U \cdot V)$, and $(U, V, U \cdot V, |U-V|)$. These inputs were then fed into a hidden network, converging at a soft-max output for classification.

\paragraph{Cross encoder to Feed Forward:}
In our second approach, we combined elements of Siamese networks and Cross Encoders. This hybrid method began by generating bi-encoder outputs for both statements and condition. These outputs were then concatenated and fed into a transformer-based cross-encoder. To maintain the distinction from statements and condition, we inserted [SEP] tokens and used attention masks when passing information through the cross encoder. We crafted these attention masks through retaining the original separate encoder masks and concatenating them for the combined input into the cross encoder. Finally, these split the outputs back into different vectors and feed both into a Neural Network for classification. This approach allowed us to leverage the contextual understanding capabilities of the cross-encoder architecture while still maintaining the separate representations needed for our classification task.

\paragraph{Final Model, AEN with KDE:}
Our final approach, Adaptable Embeddings Network (AEN) with Kernel Density Estimation (KDE), deviates from traditional mean pooling methods. For a set of attended tokens with $N$ dimensions and $M$ tokens, we transform one set of mean-pooled tokens into $N$ non-parameterized Kernel Density Functions. The other embedding is mean-pooled conventionally. We then evaluate the probability of each dimension of the mean-pooled output with respect to its corresponding density function. In essence, each dimension of the embedding is treated as its own distribution. Every token adds an additional kernel to each distribution such that we create $N$ 1D distributions with $M$ kernels per distribution. For a KDE 1D visualization refer to \ref{fig:KDE example}.
\newline This method was motivated by our hypothesis that individual embedding dimensions represent distinct attributes of the input. By translating these dimensions into probability distributions, we enable a more nuanced comparison between corresponding attributes from different inputs. This probabilistic approach allows us to evaluate how likely the attributes of one input are with respect to the distribution of attributes in another, potentially capturing more subtle relationships than traditional vector comparison methods such as Cosine Similarity. This method seemed to both show a more consistent increase in performance on test data while requiring less parameters then the other models at similar performance. 
\newline We arrived at Kernel Density Estimation after analyzing the statistical properties of embedding distributions. Using 100 sentences randomly sampled from Wikipedia, we embedded each sentence and performed Kolmogorov-Smirnov tests on their attended token distributions across corresponding embedding dimensions. The results showed a striking pattern: while the mean p-value was approximately 0.1, the median p-value fell below 0.001, indicating that most dimensions strongly rejected the hypothesis of coming from the same parametric distribution. This right-skewed distribution of p-values suggests that while some dimensions might be well-approximated by parametric distributions, the majority require non-parametric density estimation approaches like KDE rather than parametric alternatives such as GMMs. This statistical evidence guided our choice of KDE as a flexible, non-parametric method capable of handling both well-behaved and more complex distributional patterns across embedding dimensions.

\paragraph{Hyper parameters:}
Our hyperparameter exploration covered several crucial aspects of the model. We tested different pretrained encoder models to find the most effective base for our task. To address class imbalance, we experimented with various weight penalization techniques. We also investigated the impact of training data volume on model performance. The architecture of the final feed-forward neural network (FFNN) was another key area of exploration, where we tested different configurations of layers and neurons. For our final AEN model with KDE, we additionally explored which embedding to transform into a probability distribution and the type of probability density function (PDF) to apply. These hyperparameters were systematically varied to optimize our model's performance across different architectural approaches. 

\section{Results}

\subsection{Hyperparameter Comparison}

To determine the best model model we examine each hyper parameter of our models and take the highest performing option of each. We perform these experiments utilizing different Model architectures in each all other parameters are consistent.

\subsubsection{Datapoints/Batch Size}

One of the first test we ran examined results through varying access to data points and batch size. We therefore ran these experiments with the feed forward head atop both encoders.

For these experiments, we maintained consistent hyperparameters: learning rate of 2e-5, MiniLM-L6-v2 as the pre-trained encoder, 12 training epochs, class weighting of 6, and a feed-forward neural network with batch normalization and ReLU activation.

\begin{table}[h]
\centering
\begin{tabular}{|c|c|c|c|}
\hline
\textbf{Datapoints} & \textbf{Batch Size} & \textbf{F1 Score} & \textbf{Loss} \\
\hline
50k & 64 & 0.644 & 1.235 \\
250k & 128 & 0.704 & 1.203 \\
1.2m & 256 & 0.732 & 0.933 \\
2.5m & 256 & 0.752 & 0.781 \\
\hline
\end{tabular}
\caption{Model Performance on Test Data with Increasing Datapoints and Varying Batch Sizes}
\label{tab:model_performance}
\end{table}

Although each of them saw gains in F1 they all had there lowest test loss at the first epoch. The loss increased less as data points scaled to where the 2.5 million case only had an .001 increase from 6.000 to 6.001. We learned from these experiments data quantity was a key factor in model success.

\subsubsection{Pretrained Encoder}

To assess encoder performance, we compared two pre-trained models (sentence-transformers/all-MiniLM-L6-v2 and sentence-transformers/all-mpnet-base-v2) while maintaining other hyperparameters from the previous experiment (learning rate of 2e-5, 12 training epochs, class weighting of 6, and a feed-forward neural network with batch normalization and ReLU activation, trained on 2.5M datapoints).

\textit{Mpnet-base} tended to over fit quicker so we evaluate the effectiveness of these models with 2.5 million data points in there fourth epoch. 

\begin{table}[h]
    \centering
    \begin{tabular}{|l|c|c|c|}
        \hline
        \textbf{Model} & \textbf{Params} & \textbf{Test Loss} & \textbf{Test F1} \\
        \hline
        MiniLM & 22.7M & 0.654 & 0.749 \\
        mpnet-base & 109.0M & 0.800 & 0.758 \\
        \hline
    \end{tabular}
    \caption{Model Performance Comparison}
    \label{tab:model-comparison}
\end{table}

To get the best results on mpnet-based we hypothesize additional data would improve the model as observed with \textit{MiniLM} in the previous section. 

\subsubsection{Learning Rate}
To evaluate learning rate sensitivity in our Cross Encoder To Feed Forward architecture, we tested rates between 2e-6 and 2e-5 while maintaining consistent parameters (all-mpnet-base-v2 as base encoders, albert-base-v2 as cross encoder, 128 training epochs, 2.5M datapoints, and a loss weight of 6).

\begin{table}[h]
\centering
\begin{tabular}{|c|c|c|c|}
\hline
\textbf{Run} & \textbf{Epoch} & \textbf{Test Loss} & \textbf{Test F1} \\
\hline
\multirow{3}{*}{2e-5 LR} & 1 & 0.734 & 0.616 \\
& 2 & 1.237 & 0.363 \\
& 3 & 1.224 & 0.369 \\
\hline
\multirow{3}{*}{2e-6 LR} & 1 & 0.566 & 0.736 \\
& 2 & 0.596 & 0.755 \\
& 3 & 0.688 & 0.760 \\
\hline
\end{tabular}
\caption{Comparison of Test Metrics Across Training Runs}
\label{tab:test-metrics}
\end{table}

Here we find dropping the LR significantly increases performance demonstrating superior learning. Note however that loss continued to increase after epoch 2 with the decreased learning rate. 

We also tested a varied learning rate on the AEN architecture. Hyperparamters included the MiniLM, Gaussian KDE, 2.5 million datapoints, 8 epochs and a simple weight matrix to transpose the out coming probabilities from the KDE into one number so that a softmax operation could be perfomed. 

\begin{table}[h]
    \centering
    \begin{tabular}{|c|c|c|}
        \hline
        \textbf{Learning Rate} & \textbf{Loss} & \textbf{F1} \\
        \hline
        0.000002 & 0.618 & 0.637 \\
        0.00001 & 0.57 & 0.665 \\
        \hline
    \end{tabular}
    \caption{Comparison of Test Metrics Across Learning Rates}
    \label{tab:lr_comparison}
\end{table}  

Both models showed consistent improvement throughout the eight epochs, with decreasing loss and increasing F1 scores. While the higher learning rate model demonstrated superior performance, its F1 scores were relatively lower compared to other architectures. However, the AEN model's steady improvement in both loss and F1 metrics, without plateauing at the end of training, suggests robust learning rather than over-fitting. This consistent progression provides strong evidence for AEN's effectiveness.

\subsubsection{Loss Weight}
To investigate the impact of class weighting on our AEN architecture, we tested weights ranging from 1 to 6 while maintaining consistent parameters (MiniLM encoder, Gaussian KDE, 2.5M datapoints, 4 training epochs, 2-e6 learning rate and a weight matrix for KDE probability transformation). 

Because the data discrepancy in 1:6 for most models we use 6 as the loss weight to penalize positive class classification more heavily. 

\begin{table}[h]
    \centering
    \begin{tabular}{|c|c|c|}
        \hline
        \textbf{Weight} & \textbf{Loss} & \textbf{F1} \\
        \hline
        6 & 0.696 & 0.584 \\
        3 & 0.486 & 0.654 \\
        1 & 0.260 & 0.692 \\
        \hline
    \end{tabular}
    \caption{Impact of loss weights on model performance metrics}
    \label{tab:loss_weight_comparison}
\end{table}

The discrepancy in the F1 scores can be attributed to the difference in precision and recall based on the weight. The larger the weight the more precision and less recall the model has initially as shown in \ref{tab:precision-recall-comparison}.

\begin{table}[htbp]
\centering
\begin{tabular}{|c|cc|cc|}
\hline
\multirow{2}{*}{Epoch} & \multicolumn{2}{c|}{Weight = 1} & \multicolumn{2}{c|}{Weight = 6} \\
& Precision & Recall & Precision & Recall \\
\hline
1 & 0.685 & 0.475 & 0.367 & 0.929 \\
2 & 0.736 & 0.512 & 0.402 & 0.934 \\
3 & 0.661 & 0.686 & 0.408 & 0.944 \\
4 & 0.691 & 0.678 & 0.426 & 0.945 \\
\hline
\end{tabular}
\caption{Comparison of Test Precision and Recall across Epochs for Different Loss Weights}
\label{tab:precision-recall-comparison}
\end{table}

\subsubsection{KDE bandwidth}

We test two different methods of bandwidth estimation, Scott and Silverman. For all other KDE tests we use Scott exclusively. Hyperparameters included the MiniLM, Gaussian KDE, 2.5 million data points, 8 epochs and a simple weight matrix to transpose the KDE results.

\paragraph{Bandwidth Estimation Methods}
Both Scott's and Silverman's rules provide data-driven approaches for selecting the kernel bandwidth as a function of total kernels. For our implementation where we analyze each embedding dimension independently, Scott's rule is defined as:

$$h = n^{-1/5} \sigma$$

where $n$ is the sample size (number of tokens), and $\sigma$ is the standard deviation. Note that $d=1$ in our case as we perform separate univariate KDEs for each embedding dimension. Silverman's rule, also for univariate data, is given by:

$$h = (4/(3n))^{1/5} \sigma$$

Scott's rule typically produces slightly larger bandwidths than Silverman's, resulting in more conservative smoothing of the probability density estimates.

\paragraph{Performance Analysis}
We find that Scott performs better as shown in Table~\ref{tab:bandwith_comparison}. The superior performance of Scott's rule (F1: 0.637 vs 0.607) can be attributed to several factors:
\begin{itemize}
    \item Its larger bandwidths help prevent overfitting to individual token embeddings aiding generalization
    \item More robust kernel estimation across each of the 384 dimensions of our embedding space
    \item Better performance when dealing with non-Gaussian distributions in individual embedding dimensions
\end{itemize}

\paragraph{Theoretical Implications}
The performance difference between these methods reveals important characteristics of our embedding space:
\begin{itemize}
    \item While both rules are designed for univariate data, Scott's more conservative bandwidth estimation appears to better capture the underlying token distribution in each dimension
    \item The performance gap suggests our embedding distributions may be multimodal or non-Gaussian as Scott's rule does a better job of generalization
\end{itemize}

\begin{table}
    \centering
    \begin{tabular}{|l|c|c|}
        \hline
        Model & Test Loss & F1 Score \\
        \hline
        Silverman & 0.660 & 0.607 \\
        Scott & 0.618 & 0.637 \\
        \hline
    \end{tabular}
    \caption{Bandwith Performance Comparison}
    \label{tab:bandwith_comparison}
\end{table}

\subsubsection{KDE Function}
To evaluate different kernel density estimation methods, we compared three functions (Gaussian, Epanechnikov, and Triangular) while maintaining consistent parameters (MiniLM encoder, 2.5M datapoints, 8 training epochs, 2-e6 learning rate, and a weight matrix for probability transformation).

The kernel functions are defined as \cite{chung2004gaussian, epanechnikov1969nonparametric, richardson2023sharp, razakarivony2020generalizedmeanshifttriangular}:
\[
K_{\text{Gaussian}}(u) = \frac{1}{\sqrt{2\pi}}e^{-u^2/2}
\]
\[
K_{\text{Epanechnikov}}(u) = \frac{3}{4}(1-u^2)\mathbf{1}_{|u|\leq 1}
\]
\[
K_{\text{Triangular}}(u) = (1-|u|)\mathbf{1}_{|u|\leq 1}
\]

where $u = \frac{x-x_i}{h}$ for a given point $x$ and kernel center $x_i$, with bandwidth $h$. $\mathbf{1}$ represents the indicator function.

Each kernel contains distinct distribution properties:
\begin{itemize}
    \item \textbf{Gaussian}: Non-zero probability across all real numbers, though quickly approaching zero
    \item \textbf{Epanechnikov}: Zero probability beyond distance of one bandwidth unit from center
    \item \textbf{Triangular}: Linear decrease to zero at one bandwidth unit from center
\end{itemize}

\begin{table}[h]
    \centering
    \begin{tabular}{|l|c|c|}
        \hline
        \textbf{KDE Function} & \textbf{Test F1} & \textbf{Test Loss} \\
        \hline
        Gaussian & 0.637 & 0.618 \\
        Epanechnikov & 0.611 & 0.664 \\
        Triangular & 0.600 & 0.679 \\
        \hline
    \end{tabular}
    \caption{Comparison of KDE estimation functions and their performance metrics}
    \label{tab:kde_comparison}
\end{table}

We find that considering dimensions as Gaussian Probability Density Function  produces the most effective results as shown in Figure \ref{tab:kde_comparison}. We see attributed several factors potentially contributing:
\begin{itemize}
    \item The non-zero probability at all distances allows every token to contribute to the density estimate, though distant tokens have negligible impact
    \item Smooth derivatives enable more stable gradient flow during training
    \item The gradual exponential decay better captures semantic relationships between tokens compared to kernels with hard cutoffs
\end{itemize}

\subsubsection{KDE Application}

When using a KDE function in this manner one of the encoders will output a mean pooled fixed vector and the other will produce a function determined by all embedding output tokens. We test applying the KDE to both the Conditions and Statements. Parameters include MiniLM encoder, 2.5M datapoints, 4 training epochs, and a weight matrix for probability transformation. We find there is little difference and elect to apply the KDE to the statements as they have more tokens so create a more complete distribution.

\begin{table}[h]
    \centering
    \begin{tabular}{|l|c|c|}
        \hline
        \textbf{KDE transform} & \textbf{Test Loss} & \textbf{Test F1} \\
        \hline
        Caption & 0.618 & 0.637 \\
        Threshold & 0.613 & 0.636 \\
        \hline
    \end{tabular}
    \caption{Comparison of which set of output tokens are transformed into a set of Probability Density Functions}
    \label{tab:embedding_comparison}
\end{table}

We attribute the marginal performance increase to statements containing more tokens on average, allowing for more robust density estimation across each embedding dimension.

\subsubsection{Additional Hyper parameters}

We test several additional hyper parameters worth briefly discussing. 

\paragraph{Head Network Parameters}
Across all architectures we test a variety of network heads balancing parameters with over fitting. In the end we elect for a small dense network with few layers, batch norm and RELU activations.  

\paragraph{Concatenation method}
We found in our experiments that concatenation methods for the Bi-encoder feed forward and the bi-encoder/cross-encoder feed forward networks mirrored similar relative performance of the SBERT paper \cite{reimers2019sentencebertsentenceembeddingsusing}. In the end this information did not prove relevant for the AEN model.

\subsection{SLM Comparison}

To evaluate the effectiveness of our AEN model, we compared its performance against a state-of-the-art small language model, LLaMA 3.2 3B. We conducted this comparison across several key metrics: Precision, Recall, F1, and computational efficiency using 5000 additional sampled generated.

\begin{table}[htbp]
\centering
\begin{tabular}{|l|c|c|c|c|}
\hline
\textbf{Model} & \textbf{Acc.} & \textbf{Prec.} & \textbf{Rec.} & \textbf{F1} \\
\hline
AEN & \textbf{.88} & .63 & .90 & \textbf{.74} \\
\hline
\multicolumn{5}{|l|}{\textbf{LLaMA-3.2-3B (16-bit quantization)}} \\
\hline
\quad COT & .43 & .24 & .90 & .38 \\
\quad Plain & .49 & .27 & \textbf{.95} & .42 \\
\quad MultiShot & .84 & \textbf{.69} & .31 & .43 \\
\hline
\multicolumn{5}{|l|}{\textbf{Phi-3.5-mini-instruct(8-bit quantization)}} \\
\hline
\quad COT & .71 & .38 & .90 & .54 \\
\quad Plain & .65 & .34 & .88 & .49 \\
\quad MultiShot & .75 & .42 & .78 & .54 \\
\hline
\end{tabular}
\caption{Comparison of AEN and LLaMA Variants}
\label{tab:model-comparison}
\end{table}

We select the largest quantized model below 10GB of storage for each of these models to reflect low recourse environment conditions. 

COT performed poorly because it predicted the positive class incorrectly most of the time. We hypothesize the model looked for any possible reason to find a connection rather then looking for full satisfaction in the condition.

\begin{table}[htbp]
\small
\begin{tabular}{|l|r|r|c|}
\hline
\textbf{Model} & \textbf{Params} & \begin{tabular}[c]{@{}c@{}}\textbf{FLOPs}\\\textbf{/Pass}\end{tabular} & \textbf{Input} \\
\hline
AEN & 219M & 22.4B & \begin{tabular}[c]{@{}c@{}}bs=1, len=128\end{tabular} \\
\hline
LLaMA-3.2-3B & 3.2B & 360.9B & bs=1, len=128 \\
\hline
Phi-3.5-mini & 3.82B & 464.0B & bs=1, len=128 \\
\hline
\end{tabular}
\caption{Computational requirements for AEN and LLaMA-3.2-3B models. Note both encoders receive the input dimensions in the AEN. BS stands for batch size and len is the sequence length in terms of tokens.}
\label{tab:model-comparison}
\end{table}

\section{Applications}

\subsection{Edge Computing}
The computational efficiency demonstrated by AEN makes it ideal resource-constrained edge environments. Our experimental results, showing significant reductions in both parameter count and FLOPs relative to SLMs suggest viability in real-time monitoring applications.
One implementation involves positioning AEN downstream of speech-to-text systems for real-time text classification. This configuration offers several advantages:

\begin{itemize}
\item \textbf{Local Processing}: On device inference eliminates the need for continuous cloud transmission of sensitive conversational data.
\item \textbf{Conditional Data Transfer}: Binary classifications can serve as triggers for selective cloud uploads, implementing efficient data triage based on semantic criteria. 
\item \textbf{Dynamic Criterion Updates}: The separation of statement and condition encoders in our architecture allows for criterion pre-processing, reducing runtime computational requirements and parameter storage by approximately 50\%. The efficiency gain stems from the model's ability to compute and cache condition embeddings prior to deployment, requiring only the statement encoder and classification head to process incoming text at runtime (as shown in Figure \ref{fig:run}). This architectural choice effectively halves the inference-time compute expense relative to processing both inputs simultaneously. \end{itemize}

The architecture's ability to process multiple criteria simultaneously through the condition encoder further enhances an already compelling edge compute case. To illustrate this in practice, consider AEN's deployment on a wearable device interfacing with ambient audio. In this paradigm, the AEN could be seeded with multifaceted classification criteria encompassing emergency detection, information retrieval, and intentionality recognition. For instance, when a colleague expresses interest in convening and the wearable user acquiesces, the AEN could identify this concordance and flag it for a subordinate scheduling agent—exemplifying the system's capacity for semantic triage.

\subsection{Decision Trees}
AEN's binary classification architecture extends to decision tree implementations, where some traditional boolean operators could be replaced with natural language criteria. Leveraging our model's ability to evaluate semantic conditions efficiently, the resulting tree could potentially discern additional insights.

Several challenges also emerge in this application. Primary among these is optimal criterion selection - determining which natural language conditions best partition the semantic space at each decision node. We hypothesize this challenge might be addressed through language models incorporation into the training process to generate branching criteria. This approach could automate the construction of semantic decision trees while maintaining AEN's efficiency advantages. We leave the exploration of these methods to future work.

\section{Conclusion and Discussion}

\subsection{Theoretical Analysis of KDE in Embedding Comparison}

\textbf{Distribution-based Representation}: Unlike point-based measures like cosine similarity, KDE creates a continuous probability distribution for each embedding dimension. This allows for:
\begin{itemize}
\item Capturing uncertainty in embedding representations
\item Consideration of the full token distribution rather than just mean pooling
\item More nuanced comparison of semantic relationships
\end{itemize}
\textbf{Individual Dimension Treatment}: By treating each embedding dimension as its own distribution, the KDE method provides several key benefits:
\begin{itemize}
\item The model captures different scales of variation across semantic dimensions
\item Local patterns in specific semantic aspects are preserved
\item The architecture becomes more robust to outlier tokens in individual dimensions
\end{itemize}
Crucially, for an embedding of dimension $d$, KDE transforms the input into $d$ separate probability distributions. Each distribution represents the likelihood of a particular semantic feature being present. This differs fundamentally from traditional similarity measures which collapse the comparison into a single scalar value. The output can therefore be processed by additional network layers before final classification, allowing the model to learn complex relationships between semantic features.

\subsection{Data}

Data generation emerged as a primary challenge in our research. Our experimental results with varying dataset sizes (Table \ref{tab:model_performance}) suggest significant potential for performance improvements with increased data volume. The consistent gains in F1 score observed when scaling from 50k to 2.5M datapoints (0.644 to 0.752) in our classical FFNN implementation indicate that further scaling by orders of magnitude could yield additional performance improvements.

Several key areas for improvement in our data pipeline emerge:
\begin{itemize}
\item \textbf{Real-world Alignment}: The scarcity of publicly available transcript data necessitated our synthetic generation approach. Future work could benefit from partnerships providing access to authentic conversational data, improving the model's generalization to real-world scenarios.
\item \textbf{Labeling Enhancement}: While GPT-4o-mini demonstrated reasonable performance in our labeling pipeline, human evaluation revealed accuracy gaps suggesting two potential paths for improvement:
\begin{itemize}
    \item More sophisticated language models for labeling
    \item A ensemble model consensus labeling approaches
\end{itemize}

\item \textbf{Synthetic Data Filtering}: Implementation of quality control mechanisms to identify and filter highly synthetic examples could improve dataset quality while maintaining the advantages of our generation pipeline described in Section 3.1.
\end{itemize}

These improvements would likely yield superior model alignment and performance. 

\subsection{Pre-trained embedding selection}
While our experimental results demonstrate strong performance using all-mpnet-base-v2 and all-MiniLM-L6-v2 as foundation models (Table \ref{tab:model-comparison}), these approaches do not leverage recent advances in embedding techniques. State-of-the-art methods such as LLM2VEC \cite{behnamghader2024llm2veclargelanguagemodels} offer potential improvements through their enhanced natural language understanding capabilities and more sophisticated semantic representations. We opted against using LLM2VEC models since their derivation from billion-parameter language models would negate the computational efficiency advantages offered by our chosen lightweight embedding approaches.

\subsection{Fine-tuning Techniques}
Our approach could benefit from the implementation of sophisticated fine-tuning methods such as Low-Rank Adaptation (LoRA) \cite{hu2021loralowrankadaptationlarge}. LoRA could provide easier domain implementation. We opted against using LoRA due to potential minor degradations in model performance.

\subsection{Conclusion}

The AEN serves as a efficient multi-input classifier build on foundational embedding models. Data collection of device serves an integral purpose in the modern age. We created the AEN as a means of numerically understanding incoming natural language. Our contribution notably uses Kernel Density Functions as a method of comparing sentence level embeddings. Utilizing a KDE proved as effective if not more so then feeding in standard concertinaed outputs into a network head. To out knowledge, we achieved the first natural language adaptable binary classifier.

\bibliographystyle{plainnat} 
\bibliography{sources.bib}

\clearpage
\appendix

\section{Prompting Techniques}\label{app:prompts}

Below details a subset of multi shot examples we used at each stage of synthetic generation. Examples are randomly selected per datapoint. 

\subsection{Statements}

Here we detail two of the 16 different examples we used in the Multishot generation. 
\subsubsection*{Example 1: Party Planning Discussion}

\begin{quote}
\textbf{Charlotte:} Hey Sarah, I was thinking about throwing a big 4th of July party this year. Would you want to help me plan it?

\textbf{Sarah:} Yeah, that sounds awesome! I love 4th of July parties. Where were you thinking of having it?

\textbf{Charlotte:} I was thinking we could do it at my place, since I have that big backyard with plenty of space for people to hang out.

\textbf{Sarah:} Ooh perfect, your yard would be great for that. We could set up some lawn games and maybe even a little dance floor area.

\textbf{Charlotte:} I like the way you think! And of course we'll need to grill up a ton of food. Burgers, hot dogs, maybe some BBQ chicken?

\textbf{Sarah:} Definitely, it wouldn't be a 4th of July party without a big cookout. And we can ask people to bring side dishes potluck style.

\textbf{Charlotte:} Good call. Hey, what do you think about setting up a Facebook event to invite people and coordinate everything?

\textbf{Sarah:} Hmm, I'm not really a fan of using Facebook for stuff like this. I feel like it's hard to keep things organized there.

\textbf{Charlotte:} Really? I've found Facebook events pretty helpful for parties in the past. But I'm open to other ideas if you have a better suggestion!

\textbf{Sarah:} What about a shared Google Doc instead? We can make different sections for the guest list, food sign-ups, supplies we need, a schedule for the day, etc.

\textbf{Charlotte:} Okay, I can see that working well. We can share the Doc link on the actual invite.

\textbf{Sarah:} Yeah, an email invite with the Doc link is perfect. We can make it really festive and Fourth of July themed.

\textbf{Charlotte:} Sounds great. And I'll start brainstorming ideas for decorations and activities. Maybe we could even put together little welcome bags for everyone with mini flags and sparklers and stuff.

\textbf{Sarah:} Ooh I love that! Very festive. I'll add a section for welcome bag ideas to the Doc too. This is going to be such a fun party!
\end{quote}

\subsubsection*{Example 2: Vacation Planning}
\begin{quote}
\textbf{Olivia:} I've been thinking about our next vacation. How about a trip to the mountains?

\textbf{Ethan:} I don't know. The mountains can be so cold and remote. Why not a beach destination instead?

\textbf{Olivia:} I get that, but the mountains offer a peaceful retreat. We could do some hiking, enjoy the fresh air, and get away from the crowds.

\textbf{Ethan:} But what if we want some activities and nightlife? The beach has so much more to do.

\textbf{Olivia:} True, but the mountains have their own charm. Think about the cozy cabins, the beautiful sunsets, and maybe even a bit of snow. Plus, we can still find local events and things to do in the nearby town.

\textbf{Ethan:} Hmm, I hadn't thought about it that way. A cabin sounds nice. We could use this trip to really unplug and relax.

\textbf{Olivia:} Exactly! And we can always check out nearby attractions if we get bored. I'll look up some cabin rentals and send you the links.

\textbf{Ethan:} Alright, you've convinced me. Let's go for it! I'll start packing some warm clothes.

\textbf{Olivia:} Great! I'll handle the bookings and share the details with you later.

\textbf{Ethan:} I hope we can find a cabin with a fireplace. That would be perfect for chilly evenings.

\textbf{Olivia:} I'll make sure to find one. We can also plan some fun activities like skiing or snowboarding if there's enough snow.

\textbf{Ethan:} That sounds exciting! Maybe we can also find a local spa for a relaxing day.

\textbf{Olivia:} Definitely! A spa day would be a nice treat. I'll look for cabins near a good spa.

\textbf{Ethan:} This trip is shaping up to be a perfect blend of adventure and relaxation. Can't wait!

\textbf{Olivia:} Me neither. Let's finalize everything by the end of the week so we can start counting down the days.
\end{quote}

\subsection{Conditions}
These examples expand upon the conversations presented in the previous statements section. Each example demonstrates multiple instances where specific thresholds are met within natural dialogue.

\subsubsection{Example 1 Cont.}
\begin{itemize}
    \item When someone asks for planning help
    \item When someone mentions using Facebook events or Google Docs
    \item When someone suggests an actionable planning item
\end{itemize}

\subsubsection{Example 2 Cont.}
\begin{itemize}
    \item When someone discusses vacation planning
    \item When someone talks about booking accommodations
    \item When someone mentions looking up local attractions
\end{itemize}
Each conversation is designed to contain 2-4 distinct thresholds that may be triggered multiple times throughout the dialogue. 
\subsection{Labels}
Last we present examples showing how the model evaluates individual statements against a set of conditions.

\subsection*{Example 1: Using Digital Assistants}
\textbf{Input Statement:}
\begin{quote}
``Jamie: I use Siri all the time to set reminders and check the weather. It's so convenient.''
\end{quote}

\textbf{Conditions Tested:}
\begin{enumerate}
    \item When someone orders food using a food delivery app
    \item When someone expresses excitement about improving workflow and productivity
    \item When someone discusses the impact of AI on daily activities
    \item When someone talks about using AI-powered assistants
    \item When someone suggests joining a fitness challenge using a fitness tracking app
\end{enumerate}

\textbf{Classification Results:} [0, 0, 1, 1, 0]

\subsection*{Example 2: Scientific Observation}
\textbf{Input Statement:}
\begin{quote}
``David: Speaking of observation, have you ever used a time-lapse camera to record the feeding behavior of the octopuses?''
\end{quote}

\textbf{Conditions Tested:}
\begin{enumerate}
    \item When someone offers assistance with art supplies
    \item When someone reflects on emotions tied to a place or experience
    \item When someone plans a scientific observation project
    \item When someone proposes recording footage for research purposes
    \item When someone talks about analyzing animal behavior
\end{enumerate}

\textbf{Classification Results:} [0, 0, 0, 1, 1]

\end{document}